\pdfoutput=1

\documentclass[11pt]{article}

\usepackage[final]{acl}

\usepackage{times}
\usepackage{latexsym}

\usepackage[T1]{fontenc}

\usepackage[utf8]{inputenc}
\usepackage{enumitem}

\usepackage{microtype}

\usepackage{inconsolata}

\usepackage{graphicx}

\usepackage{float}
\usepackage{booktabs}
\usepackage{array}
\usepackage{multirow}
\usepackage{caption}
\usepackage{tcolorbox}
\usepackage{xcolor}
\usepackage{xcolor}
\usepackage{multicol}
\usepackage{pifont}
\usepackage{changepage}

\definecolor{lightgreen}{RGB}{145, 204, 117}
\definecolor{darkblue}{RGB}{0, 0, 255}
\newtcolorbox{promptbox}[2][Prompt]{
colback=blue!5!white,  
arc=5pt,
boxrule=0.5pt,
fonttitle=\bfseries,
title=#1,
before upper={\small}, fontupper=\fontfamily{ptm}\selectfont,
colframe=#2,
width=\textwidth,           
}

\newtcolorbox{simpledashbox}[2]{
colback=#1!10!white,   
colframe=#1!70!black,  
arc=5pt,              
boxrule=0.5pt,
width=\linewidth,
before skip=5pt, after skip=5pt
}
\usepackage{tcolorbox}
\usepackage{xcolor}

\usepackage{amsmath,amsfonts,bm}

\def\eqref#1{equation~\ref{#1}}

\def\1{\bm{1}}

\DeclareMathAlphabet{\mathsfit}{\encodingdefault}{\sfdefault}{m}{sl}
\SetMathAlphabet{\mathsfit}{bold}{\encodingdefault}{\sfdefault}{bx}{n}

\newcommand{\cD}{\mathcal{D}}

\newcommand{\PP}{\mathbb{P}}

\newcommand{\E}{\mathbb{E}}

\newcommand{\KL}{D_{\mathrm{KL}}}

\usepackage{url}

\usepackage{hyperref}
\usepackage{cleveref}

\title{From Lists to Emojis: How Format Bias Affects Model Alignment}

\author{Xuanchang Zhang\thanks{Equally contribution; more junior authors listed earlier and the full authorship contribution statements are provided in the Appendix. Correspondence to: Wei Xiong $<$wx13@illinois.edu$>$} $^1$, \, Wei Xiong$^{1*}$, \, Lichang Chen$^{2*}$, 
\\ \textbf{Tianyi Zhou$^2$, \, Heng Huang$^{2}$, \, Tong Zhang$^{1}$}\\ 
$^1$University of Illinois Urbana-Champaign \\
$^2$University of Maryland College Park\\
}

\begin{document}
\maketitle
\begin{abstract}
In this paper, we study format biases in reinforcement learning from human feedback (RLHF). We observe that many widely-used preference models—including human evaluators, GPT-4, and top-ranking models on the RewardBench benchmark—exhibit strong biases towards specific format patterns, such as lists, links, bold text, and emojis. Furthermore, large language models (LLMs) can exploit these biases to achieve higher rankings on popular benchmarks like AlpacaEval and LMSYS Chatbot Arena.

One notable example is verbosity bias, where current preference models favor longer responses that appear more comprehensive, even when their quality is equal to or lower than shorter responses.
However, format biases beyond verbosity remain largely underexplored.
In this work, we extend the study of biases in preference learning beyond the commonly recognized length bias, offering a comprehensive analysis of a wider range of format biases. Additionally, we show that with a small amount of biased data (less than 1\%), we can inject significant bias into the reward model. Moreover, these format biases can also be easily exploited by downstream alignment algorithms, such as \texttt{best-of-n sampling} and online iterative \texttt{DPO}, as it is usually easier to manipulate the format than to improve the quality of responses. Our findings emphasize the need to disentangle format and content both for designing alignment algorithms and evaluating models.

\end{abstract}

\section{Introduction}

\textit{Reinforcement Learning from Human Feedback} (RLHF) \citep{ziegler2019fine, ouyang2022training, bai2022training} has become a critical technique in the LLM training pipeline to align the outputs with human preferences. 
This approach has contributed to the tremendous successes of various state-of-the-art LLMs, such as Gemini \citep{team2023gemini}, Chat-GPT \citep{OpenAI2023GPT4TR}, and Claude \citep{Anthropic@claude}, by making the generated responses helpful, harmless, honest and controllable \citep{ouyang2022training, bai2022training}. 
Despite its effectiveness, RLHF also faces challenges. 
One notable issue is that policies optimized via RLHF tend to develop biases, leading them to ``game'' the reward model~(RM) 
\textit{i.e.}, the policy receives a high reward from RMs but does not achieve intended objectives.
This behavior often results in performance degradation \citep{bai2022training, OpenAI2023GPT4TR}, a phenomenon widely referred to as reward hackings~\citep{anthropic2024rewardtampering, alex2024warp}. 

One prevalent form of reward hacking is the verbosity issue \citep{chen2024odin, length-bias-on-alpaca-eval, singhal2023long}, where the policy generates unnecessarily lengthy responses in an attempt to appear more helpful or comprehensive. 
However, many other types of bias patterns beyond the verbosity remain largely under-explored, especially in the current most popular instruction-following (IF) benchmarks. 
Recent studies reveal LLM evaluators exhibit inherent biases including self-preference for their own generations \citep{panickssery2024llmevaluatorsrecognizefavor} and style-over-substance judgments \citep{wu2023stylesubstanceevaluationbiases}, which further complicate evaluation validity.
In this study, we are dedicated to answering the following question:
\begin{center}
    \textit{On the current popular IF benchmarks, do the scores truly reflect models' capabilities? }
\end{center}
We first summarize the response statistics generated by the top-ranking models on the AlpacaEval Leaderboard \citep{alpaca_eval}. 
The results are presented in Table~\ref{tab:dataset_stat}.
We observe that all the models \textbf{strongly prefer} generated responses that include bold text, lists, and exclamation patterns. 
Additionally, the Gemma models \citep{team2024gemma, zhou2024wpo} tend to incorporate more emojis in their responses compared to Llama \citep{meta_llama3} and Mistral \citep{jiang2023mistral, liu2024iterative}.
Although the proportion of responses containing links is less significant, we notice that AlpacaEval's judge, GPT-4, tends to favor responses with links when they are present.
Due to space constraints, a more comprehensive study of these biases in existing preference models, are deferred to Section~\ref{sec:observation}.




The underlying cause of these reward hacking phenomena lies in the imperfections of the reward models used in RLHF.
These models often suffer from approximation errors and lack sufficient representativeness due to the limited diversity and number of human labelers.
For instance, in a typical RLHF setup, the target preference is that of real-world users, yet we rely on a small group of human labelers to annotate the responses generated by LLMs.
During the policy improvement stage, a proxy reward model, constructed from these annotated responses, is ultimately used to provide the learning signal.
As a result, achieving a higher proxy reward does not always correlate with better performance when evaluated by real-world users.
Even though one believes that some format biases are intrinsic to human preference, they can still cause problems in downstream alignment tasks if not explicitly addressed through regularization or specialized algorithm designs.
This is because the most popular RLHF algorithms such as \texttt{RAFT} \citep{dong2023raft}, online iterative \texttt{DPO} \citep{rafailov2023direct, xiong2024iterative}, and \texttt{PPO} \citep{schulman2017proximal} run in an on-policy (LLMs learn from the self-generated responses) and online manner (the reward continuously labels the self-generated responses throughout the training process).
These format biases are much easier to exploit by the LLMs in pursuit of a higher reward compared to improving the content quality.
Consequently, these biases can be significantly amplified in the online and on-policy training pipeline.

In this work, we extend the study of biases in preference learning beyond the commonly explored length bias, presenting a comprehensive analysis of a broader class of pattern biases. Our findings emphasize the importance of disentangling format and content in both the design of alignment algorithms and model evaluations. We summarize our contributions as follows.

\begin{enumerate}[nosep]
    \item \textbf{Format biases widely exist in human, GPT-4, and open-source preference models.} We identify several common format biases and demonstrate that both human evaluators, GPT-4, and a range of open-source preference models exhibit strong biases toward these formats.
    \item \textbf{Existing high-ranking models on the public leaderboard exhibit strong format biases.} We show that many high-ranking models on public leaderboards like AlpcaEval \citep{alpaca_eval} and MT-bench \citep{zheng2023judging}, tend to exploit evaluator biases by generating responses with specific biased formats.
    \item \textbf{Reward modeling can be easily attacked by a small amount of biased data (less than 1\%) and leads to significant format biases in downstream alignment tasks.} We show that introducing less than 1\% biased data into a debiased preference dataset can cause the resulting reward models to develop substantial format biases. These biases are easily exploited by policy models during downstream alignment, especially in online algorithms like iterative \texttt{DPO} \citep{rafailov2023direct, xiong2024iterative} and \texttt{PPO} \citep{schulman2017proximal}.
    \item \textbf{Passive data filtering is not sufficient for reward model debiasing.} We find that passive data filtering is inadequate, as it removes a substantial portion of the training data and leads to inferior model capacity. We also explores a two-head reward construction approach to debias reward models and addressed the signal sparsity issue for specific formats.
    \end{enumerate}

\begin{table*}[t]
\small
\centering 
\setlength{\tabcolsep}{2pt}
\begin{tabular}{lcccccccccc}
\toprule
\multirow{2}{*}{\textbf{Dataset}} & \multirow{2}{*}{\textbf{Num}} & \multirow{2}{*}{\textbf{Labeller}} & \multirow{2}{*}{\textbf{Type}}  & \multirow{2}{*}{\textbf{Length}} & \multicolumn{6}{c}{\textbf{Ratio of Responses with Pattern (\%)}} \\
&&&&&  \textbf{Bold} & \textbf{List} & \textbf{Emoji} & \textbf{Exclamation} & \textbf{Link} & \textbf{Affirmative} \\
\midrule
\multirow{2}{*}{RLHFlow-Preference} & \multirow{2}{*}{700000} & \multirow{2}{*}{Mixture} &
Preferred & 167.30 & 2.40 & 14.99 & 0.49 & 15.41 & 2.15 & 5.58 \\
&&& Unpreferred & 140.01 & 1.74 & 10.77 & 0.61 & 15.93 & 1.89 & 5.19 \\
\midrule
\multirow{2}{*}{LMSYS-Arena} & \multirow{2}{*}{49865} & \multirow{2}{*}{Human} &
Preferred & 191.64 & 7.61 & 38.84 & 0.73 & 18.52 & 1.20 & 7.30 \\
&&& Unpreferred & 159.77 & 4.54 & 31.67 & 0.62 & 15.77 & 1.08 & 6.45 \\
\midrule
\multirow{2}{*}{AlpacaEval} & \multirow{2}{*}{169927} & \multirow{2}{*}{GPT-4}  &
Preferred & 325.43 & 42.76 & 61.73 & 1.99 & 23.34 & 1.58 & 10.08 \\
&&& Unpreferred & 239.36 & 16.78 & 49.96 & 1.40 & 21.84 & 1.32 & 9.11 \\
\midrule
\multirow{2}{*}{\href{https://huggingface.co/datasets/HuggingFaceH4/ultrafeedback_binarized}{Ultrafeedback-binarized}} & \multirow{2}{*}{61135} & \multirow{2}{*}{GPT-4} &
Preferred & 198.34 & 3.15 & 34.52 & 1.08 & 22.50 & 3.17 & 10.94 \\
&&& Unpreferred & 175.48 & 2.62 & 27.88 & 1.00 & 23.96 & 2.32 & 11.73 \\
\bottomrule
\end{tabular}
\caption{\footnotesize Statistics summarization of different preference datasets. We compute the proportions of samples with certain formats in both the preferred and unpreferred responses. For length, we compute the number of words of the responses. 
}
\label{tab:dataset_stat}
\end{table*}

\begin{table*}[t]
\centering 
\small
\setlength{\tabcolsep}{2pt}
\begin{tabular}{lccccccc} 
\toprule 
\multirow{2}{*}{\textbf{Model}} & 
\multirow{2}{*}{\textbf{Type}} & \multicolumn{6}{c}{\textbf{Win-Rate (\%)}} \\
& & Bold & List & Emoji & Exclamation & Link & Affirmative \\
\midrule 
GPT-4 Turbo & LLM-as-a-Judge & 89.5 & 75.75 & 86.75 & 80.5 & 87.25 & 88.75 \\
\midrule 
\href{https://huggingface.co/RLHFlow/ArmoRM-Llama3-8B-v0.1}{ArmoRM-Llama3-8B-v0.1} & Multi-head RM & 98 & 50.5 & 55 & 34.5 & 27 & 28.5 \\
\midrule 
\href{https://huggingface.co/RLHFlow/pair-preference-model-LLaMA3-8B}{Pairwise-model-Llama-3-8B} & Pairwise PM & 97 & 93.5 & 70.5 & 64.25 & 84.75 & 47.75 \\
\midrule
\href{https://huggingface.co/sfairXC/FsfairX-LLaMA3-RM-v0.1}{FsfairX-Llama-3-8B-v0.1} & BT RM & 95.5 & 68.5 & 15 & 28.5 & 64.5 & 59.5 \\
\midrule
\href{https://huggingface.co/Skywork/Skywork-Critic-Llama-3.1-8B}{Skywork-Critic-Llama-3.1-8B} & Generative Model & 99.25 & 88.75 & 97.25 & 77.75 & 75 & 85 \\
\midrule
\href{https://huggingface.co/HuggingFaceH4/zephyr-7b-beta}{Zephyr-Beta-Mistral-7B} & DPO Model & 37.5 & 50 & 26.5 & 72 & 58 & 21 \\
\midrule
\href{https://huggingface.co/NCSOFT/Llama-3-OffsetBias-RM-8B}{OffsetBias-RM-Llama-3-8B} & BT RM & 77.5 & 84 & 28 & 38 & 62 & 30.5 \\
\bottomrule 
\end{tabular}

\caption{\footnotesize Evaluation results of the pattern bias of five popular preference models. This table displays the adjusted win rates, accounting for tied conditions by allocating half to the win rate. Detailed proportions of wins, ties, and losses for each model are presented in Table~\ref{detailed results of preferecence models}. For the DPO model \href{https://huggingface.co/HuggingFaceH4/zephyr-7b-beta}{Zephyr-Beta-Mistral-7B}, the reward is computed as $ \log \frac{\pi(a|x)}{\pi_0 (a|x)}$, where $\pi$ represents the model itself and $\pi_0$ represents its base model \href{https://huggingface.co/HuggingFaceH4/mistral-7b-sft-beta}{Zephyr-Beta-Mistral-7B}.}
\label{tab:reward_model_evaluation}
\end{table*}

\section{The Pattern Bias in Preference Learning} \label{sec:observation}

In this section, we summarize some representative format biases and show that these biases widely exist in human, GPT-4, and other open-source preference models. 

\subsection{Pattern Statistics in Preference Datasets and Benchmarks}


\paragraph{Methods and datasets.} To identify the format biases existing in the preference models, we select several representative preference datasets with different prompt distributions, response generators, and preference models. 
\begin{itemize}
    \item \textbf{RLHFlow-Preference-700K}\footnote{\url{https://huggingface.co/datasets/hendrydong/preference_700K}} \citep{dong2024rlhf} consists of 700K preference pairs and has been used to train a series of strong open-source reward models or pairwise preference models. This dataset is a cleaned version of many open-source preference datasets, including HH-RLHF \citep{bai2022training}, SHP \citep{pmlr-v162-ethayarajh22a}, HelpSteer \citep{wang2023helpsteer}, UltraFeedback \citep{cui2023ultrafeedback}, UltraInteract \citep{yuan2024advancing}, Distilable-Capybara, PKU-SafeRLHF \cite{ji2024beavertails}, and Distilabel-Orca \citep{OpenOrca}. 
    \item \textbf{LMSYS-Arena-55K }\citep{chiang2024chatbot}, which consists of 55K real-world conversations between users and $>70$ state-of-the-art LLMs, as well as the preference signals from the users. This dataset is prepared for the Kaggle competition of predicting human preference on Chatbot Arena battles.
    \item \textbf{AlpacaEval} \citep{alpaca_eval} is a benchmark to test models' IF capabilities with 805 test prompts. 
    The generated responses from the model are compared to the responses from GPT-4, where the GPT-4 is also used as the judge. We collect the data from 258 tested LLMs to form this dataset.
    \item \textbf{UltraFeedback} \citep{cui2023ultrafeedback} consists of 64k prompts from diverse resources and the authors generate 4 responses per prompt using 4 different LLMs. 
    The preference is labeled by the LLM-as-a-judge with GPT-4, based on a fine-grained annotation instruction. 
    Specifically, the instruction contains 4 different aspects, including instruction-following, truthfulness, honesty, and helpfulness.
\end{itemize}
For all datasets, samples are processed into the standard format $(x, a^w, a^l)$, where $a^w$ is preferred over $a^l$ according to the preference model. 
We would like to compare the proportions of samples with certain formats in both the preferred and unpreferred responses. 
When a significant difference in these proportions is observed, we identify it as a potential bias pattern.

\paragraph{Patterns.} We identify seven distinct patterns within responses: length, emoji, bold, exclamation, list, link, and affirmative. 
The examples of these patterns are shown in Appendix~\ref{sec:examples of pattern bias}.

\paragraph{Results.} We summarize the main statistics in Table~\ref{tab:dataset_stat} and outline the main findings as follows:
\begin{enumerate}[nosep]
    \item Both GPT-4 and humans exhibit a preference for longer sentences, bold, lists, exclamation marks, and an affirmative tone. However, GPT-4's preference for these elements is typically stronger than that of humans.
    \item GPT-4 also has a preference for emojis and hyperlinks, which humans do not share.
    \item  The biases in the UltraFeedback dataset are less pronounced. This may be because UltraFeedback explicitly assesses context-related qualities such as helpfulness, instruction-following, honesty, and truthfulness—factors that are less directly related to format and determine the final preference label by the average of these scores. In contrast, other datasets rely on general preference judgments without first asking for the fine-grained scores. 
\end{enumerate}

\subsection{Bias Evaluation of Preference Model}
In this subsection, we create a collection of evaluation datasets to examine the format biases of GPT-4 and open-source preference models. Specifically, we include representative open-source BT reward model \citep{bradley1952rank, dong2024rlhf}, pairwise preference model \citep{zhao2023slic, liu2023statistical, ye2024theoretical}, multi-head reward model \citep{wang2024interpretable}, generative critic model \citep{skyworkcritic2024} and the reward model of DPO model \citep{tunstall2023zephyr} in our experiments. 

\paragraph{Evaluation dataset for specific patterns.} We generate responses using prompts from the LMSYS-Arena dataset~\citep{chiang2024chatbot}. For each prompt, we generate responses from various models, including Meta-Llama-3-8B-Instruct, Meta-Llama-3.1-8B-Instruct, Mistral-7B-Instruct-v0.2, Gemma-2-9B-It, Qwen1.5-7B-Chat, Qwen2-7B-Instruct, and ChatGLM3-6B. We then select responses that exhibit the specified pattern and remove this pattern in the responses to obtain preference pairs containing two responses that are identical except for this pattern.
For each pattern, we generate 200 pairs for evaluation.

\paragraph{Evaluation results.} For each preference model, we first evaluate the responses with and without the pattern collected previously. 
We then compute the win rate of responses with the pattern against those without it. 
For an unbiased reward model, the win rate should be close to 50\%, indicating that the model judges two responses, which are identical except for the pattern, as equally favorable.
The results are summarized in Table~\ref{tab:reward_model_evaluation}, with key observations outlined below: 
\begin{enumerate}[nosep]
\item GPT-4 displays strong bias across all patterns, which aligns with our findings of pattern bias in the GPT-4 annotated preference datasets in Table~\ref{tab:dataset_stat}.  
\item Pairwise-model-Llama-3-8B and Skywork-Critic-Llama-3.1-8B also exhibit bias across all patterns.
\item The FsfairX-Llama-3-8B-v0.1 model, trained on the RLHFlow-Preference dataset, shows bias in all patterns except for emojis and exclamations, consistent with the bias statistics results of its training dataset highlighted in Table~\ref{tab:dataset_stat}. 
\item ArmoRM-Llama3-8B-v0.1, which explicitly considers multiple context-related attributes when evaluating responses, reduces format bias by focusing on aspects less directly related to formatting. However, ArmoRM still shows a strong bias toward the bold pattern.  
\item The Zephyr-Beta-Mistral-7B model, trained on the Ultrafeedback\_binarized dataset, only shows bias in exclamation and link patterns, consistent with the bias statistics results of its training dataset highlighted in Table~\ref{tab:dataset_stat}. 
\item In particular, we notice that although \citet{park2024offsetbiasleveragingdebiaseddata} leverages debiased data to mitigate several implicit biases in the reward model FsfairX-Llama-3-8B-v0.1 and obtain a new debiased reward model OffsetBias-RM-Llama-3-8B, the debiased model still shows considerable bias toward bold and list patterns because they do not explicitly consider the format bias. Moreover, the bias in list format becomes more severe after finetuning.
\end{enumerate}

\subsection{Pattern Bias and Preference Flipping}
\citet{wu2023stylesubstanceevaluationbiases} demonstrated that answers containing factual errors are sometimes rated more favorably than those that are too brief or contain grammatical mistakes. In our experiments, we observed instances where stylistic enhancements, such as bold formatting, led GPT-4 to prefer responses with inferior content over those with superior content but less polished style.

For detailed examples illustrating this phenomenon, please refer to the Appendix~\ref{flipping}.

\section{Pattern Bias Transfer in Downstream Reward Modeling and Alignment}
In Section~\ref{sec:observation}, we show that the format biases widely exist in the existing preference model. We now conduct controllable experiments to study how these biases transfer from preference data to the reward model, and further to the downstream RLHF-aligned model. For simplicity, we focus on the bold pattern and list pattern.

\subsection{A Small Amount of Data Can Lead to Significant Reward Model Bias}
\label{sec:rm attack}

\paragraph{Base dataset.} We begin by selecting two responses for each prompt from the UltraFeedback dataset \citep{cui2023ultrafeedback}, where each prompt has multiple responses generated by different models along with corresponding scores. To construct our dataset, we adopt a pairwise comparison approach, selecting response pairs where the difference in scores exceeds a predefined threshold of 1.0. After applying this filtering approach, we obtain a preference dataset containing 71.6K pairs.

\paragraph{Attacking dataset.} To inject bias into the base dataset, we generate responses following the process described in Section~\ref{sec:observation}. We label responses with the target pattern as the preferred ones and designate augmented responses as the unpreferred ones for reward model training. When injecting both bold and list biases simultaneously, we use more data containing the list pattern, as we observe that the bold pattern bias tends to dominate if both patterns are injected in equal proportions.

\paragraph{Model, and parameter.} We use the Llama-3-8B-it~\citep{llama3modelcard} as our base model of the reward model and train on the mixture of the base dataset and the attacking dataset. We also use the evaluation datasets to evaluate the biases in the reward model. For all the experiments, we follow the hyper-parameter in \citet{dong2023raft} to train the model for 1 epoch with a max length of 3096 tokens. We use a global batch size of 128 with deepspeed stage 3. We mainly search the hyper-parameter learning rate in \{5e-7, 1e-6, 5e-6\} and choose the best one (1e-6). 

\paragraph{Results.} We summarize the main results in Table~\ref{tab:results}. First, we observe that removing all training samples containing the specific pattern does not completely eliminate pattern bias as the reward model trained on the baseline dataset (a filtered version of UltraFeedback) still shows a significant bias toward the bold pattern. Additionally, even with a very small amount of biased data (less than 1\%), we can inject bias into the final model. For example, the reward model trained on the baseline dataset achieves a 51\% win rate for the list pattern, but when trained on a mixture of the baseline dataset and just 0.7\% list-augmenting data, the model's preference for lists increases to 77.5\%.

\begin{table}[t]
\centering
\small
\begin{tabular}{lccc}
\toprule
\multirow{2}{*}{\textbf{Training Dataset}}  & \multicolumn{2}{c}{\textbf{Win-rate (\%)}} \\ & \textbf{Bold} & \textbf{List} \\ 
\midrule
Baseline & 57.5 & 51.0 \\ \midrule
Baseline + 0.14\% Bold & 61.0 & - \\ \midrule
Baseline + 0.35\% Bold & 66.0 &  -\\ \midrule
Baseline + 0.70\% Bold & 88.0 & -\\ \midrule
Baseline + 0.14\% List & - & 71.5 \\ \midrule
Baseline + 0.35\% List &- & 74.0 \\ \midrule
Baseline + 0.70\% List &- & 77.5 \\ \midrule
Baseline + 1.40\% List &- & 79.5 \\\midrule
Baseline + 0.70\% Bold + 1.40\% List &83.0 & 80.0 \\
\bottomrule
\end{tabular}
\caption{\footnotesize The results of pattern biases in the reward modeling when we introduce biased data. The baseline dataset is a filtered version of UltraFeedback, where we delete all the samples containing the specific patterns. Then, we combine the baseline dataset with a small preference dataset where the preferred response contains the pattern while the unpreferred response does not.}
\label{tab:results}
\end{table}

\subsection{Downstream Alignment Task: DPO and PPO Training} 

In this section, we apply the obtained reward models in the downstream alignment tasks. Specifically, we consider the baseline reward model, the list-biased reward model (baseline dataset + 1.40\% list-augmented data), the bold-biased reward model (baseline dataset + 1.40\% bold-augmented data), and the reward model that is attacked by both 0.70\% bold-augmented data and 1.40\% list-augmented data. We study three widely used algorithms, including the offline \texttt{DPO} \citep{rafailov2023direct}, and its online variant \citet{xiong2024iterative, tajwar2024preference}, as well as the online PPO \citep{schulman2017proximal}. In comparison, the online algorithms generate responses from the current policy model, ask an external preference model (the reward model in our case) to annotate the responses, and train on the resulting preference dataset. We also study the inference-time best-of-n (BoN) sampling \citep{nakano2021webgpt}, whose results are deferred to Appendix~\ref{sec:bon}.

\paragraph{Training setup.} We train the \texttt{DPO} in an on-policy manner, following the approach of \citet{dong2024rlhf, pace2024west, snorkelai@pair}. For each prompt, we use the SFT model to sample five responses and then construct pairs using the best and the worst responses based on the ranking. We train the \texttt{DPO} for 2 epochs with a global batch size of 128 and a learning rate of 5e-7. The maximal number of tokens is 3096 and the KL coefficient is 0.1. For \texttt{PPO} training, we mainly follow the recipe in \citet{hu2024openrlhf} to use a rollout batch size of 1024, a global training batch size of 128, and set the learning rates of actor and critic as 5e-7, and 9e-6, respectively. The initial KL coefficient is 0.01 and we train on the collected data for 1 epoch. We test the resulting policy models using the AlpacaEval \citep{alpaca_eval} prompt set and the test set split from the UltraFeedback dataset. Generations are conducted with a temperature of 1.0 and a maximum token limit of 2048.

\begin{figure}[ht]
\centering
\includegraphics[width=0.98\linewidth]
{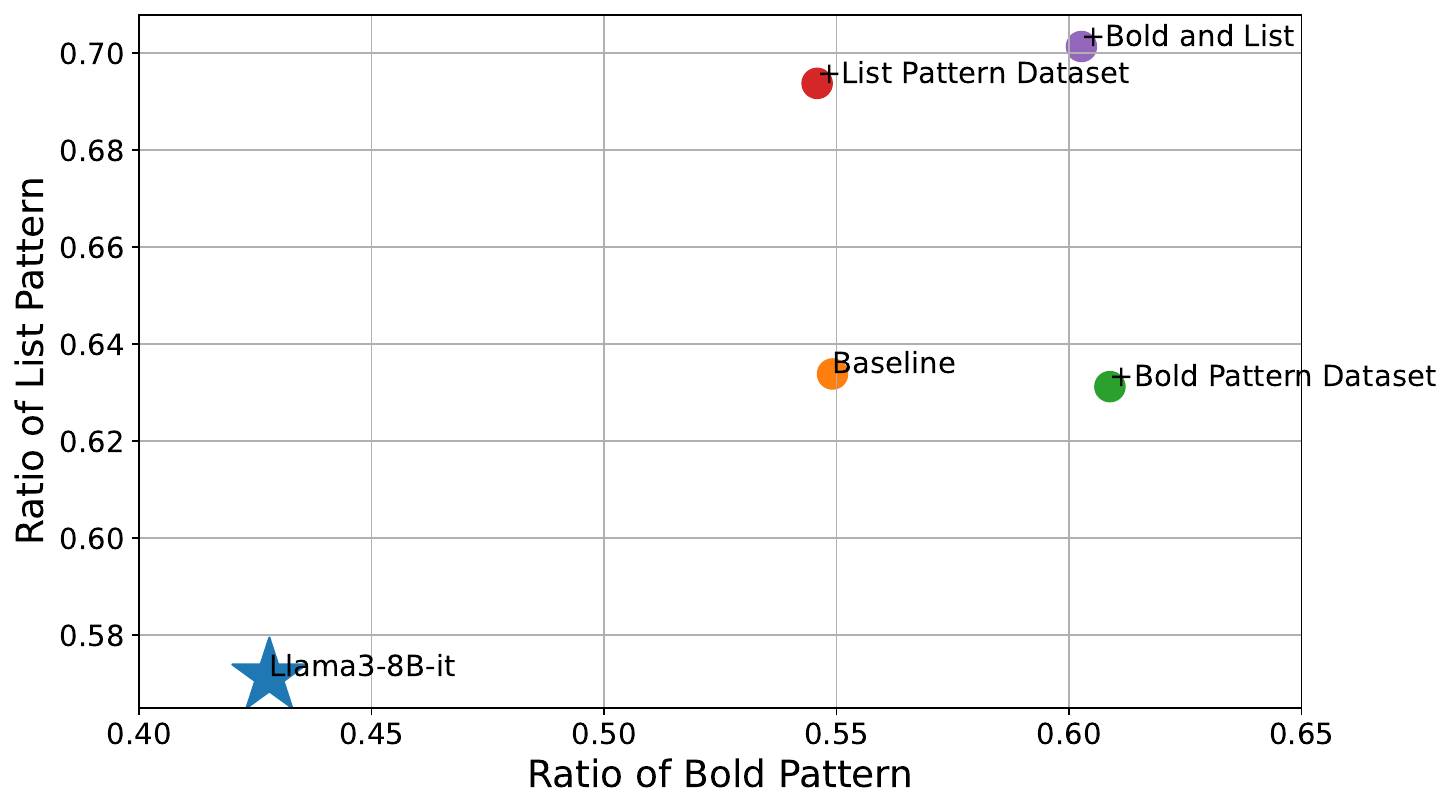}
    \caption{ \footnotesize
    We use the Llama-3-8B-it model as the base model and use the prompts from UltraFeedback to conduct \texttt{DPO} training. For bias evaluation, we generate one response for each question in AlpacaEval dataset and calculate the ratio of responses with bold/list pattern. While the base model already favors the bold and list patterns, the \texttt{DPO} training with the biased reward further amplifies these biases.
    }
    \label{fig:dpo_result1}
\end{figure}

\paragraph{Main results of offline \texttt{DPO}.} In our first experiment, we initialize with the Llama-3-8B-it model and apply different reward models from the previous section to label 59K$\times$5 on-policy samples. The results are summarized in Figure~\ref{fig:dpo_result1}. We observe that the base model, Llama-3-8B-it, already shows a strong preference for generating responses with both bold and list patterns. However, \texttt{DPO} further amplifies these biases, as all four new models shift toward the upper-right corner compared to the base model in the figure. Notably, the baseline reward model also contributes to this bias amplification, indicating that simply removing training data with the pattern in reward modeling is insufficient to fully address the issue. Additionally, the extent of bias amplification depends on the specific biases in the reward models. For example, the reward model trained with additional bold-pattern data results in a stronger bias toward bold patterns, while the one trained with both bold and list patterns amplifies both biases compared to the baseline reward model.

\paragraph{Main results of online iterative \texttt{DPO} and \texttt{PPO}.} In addition to the offline \texttt{DPO}, we also study the behavior of online algorithms including the iterative \texttt{DPO} and \texttt{PPO}. We focus on the reward model trained with both bold-augmented and list-augmented data and select Llama-3-8B-SFT from \citet{dong2024rlhf} as our base model because its biases are much lighter compared to Llama-3-8B-it, making it more suitable for illustrating the concept. To ensure a fair comparison, we fix the number of queries to the reward model for both online iterative and offline learning. Specifically, we split the UltraFeedback prompt set into three training sets, each containing 20K prompts, and a test set of 2K prompts. For offline learning, we use the entire training set at once, while in online iterative learning, we use the training sets iteratively. We also update the reference model at each iteration as done in \citet{snorkelai@pair, xiong2024building}, which achieves better in-domain performance. We also monitor the training progress of our model using the widely adopted AlpacaEval2 benchmark \citep{alpaca_eval} to ensure that our experiments reflect typical practices in online iterative \texttt{DPO} and \texttt{PPO} training. See Appendix~\ref{sec:ppo_dpo} for details.

\begin{figure}[t]
    \centering
\includegraphics[width=0.48\linewidth]{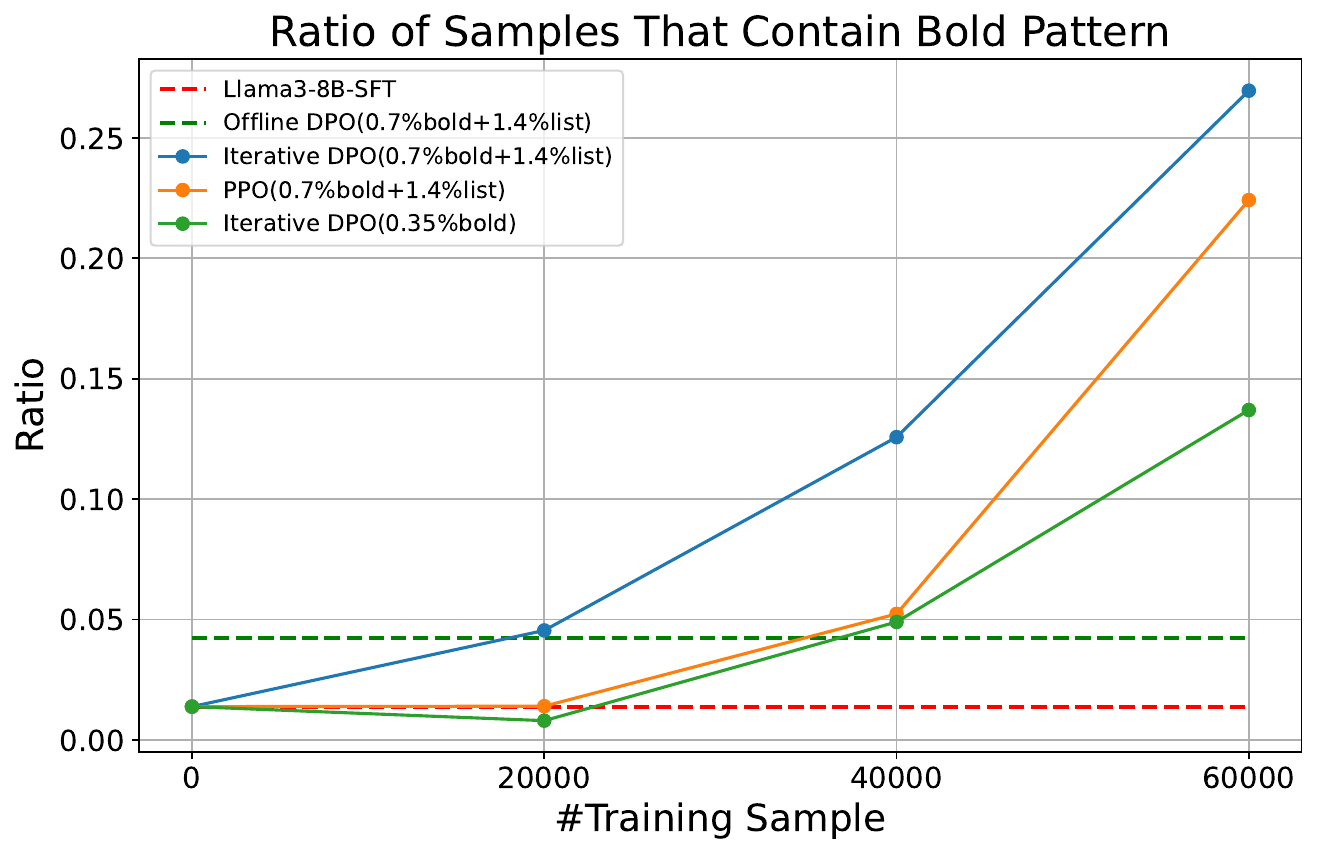}
\includegraphics[width=0.48\linewidth]{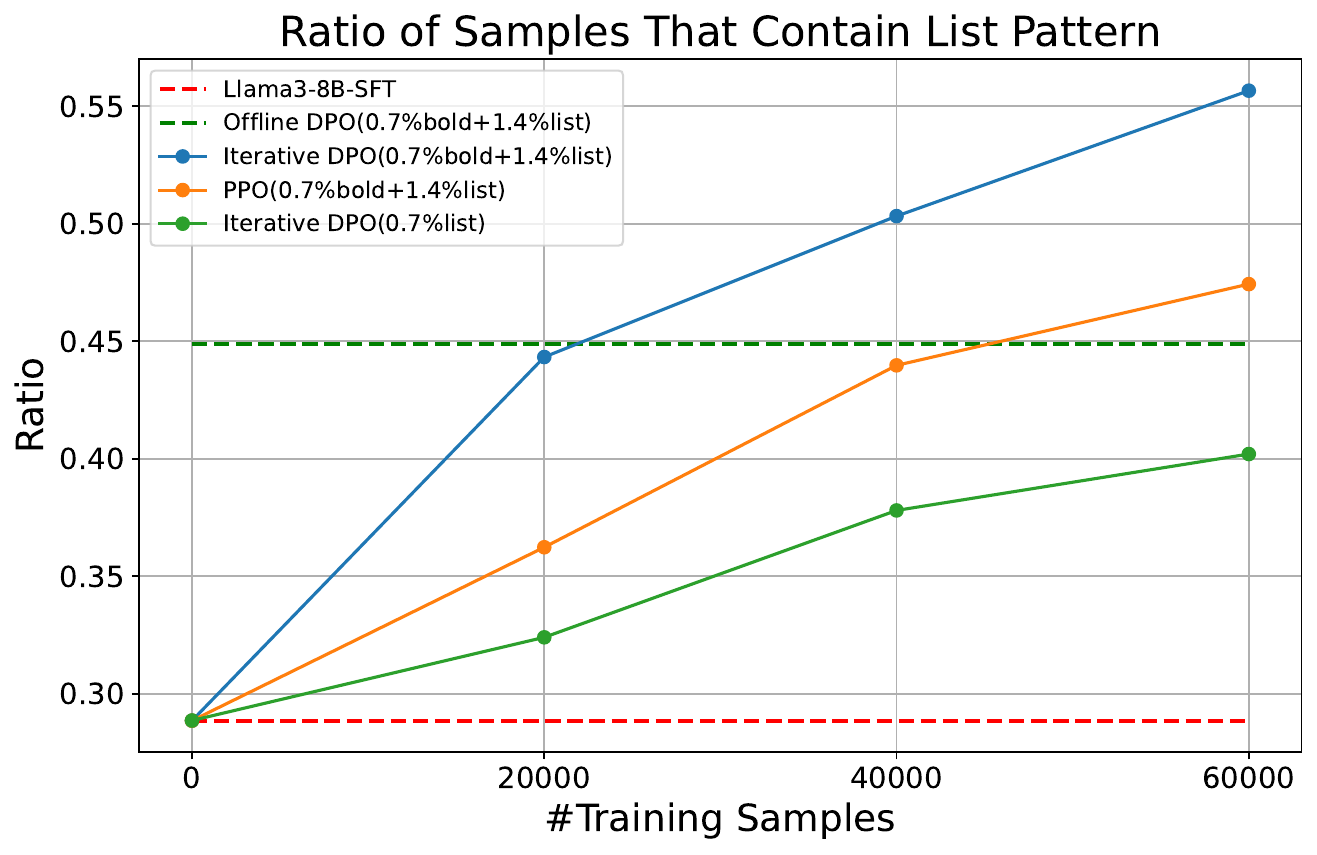}
    \caption{\footnotesize The test results on the AlpacaEval and split test set of UltraFeedback. 
    We use the Llama-3-8B-SFT model as the base model and use the prompts from UltraFeedback to conduct iterative \texttt{DPO} and \texttt{PPO} training, where the reward model is used to annotate the on-policy samples from the model. Three reward models, trained in Section~\ref{sec:rm attack} with varying sizes of attacking data, are used in this process. The offline \texttt{DPO} is a special case where we use all the prompts in one iteration. The online iterative \texttt{DPO} and \texttt{PPO} significantly amplify the pattern bias throughout training.
    }
    \label{fig:dpo_result2}
\end{figure}

The results are summarized in Figure~\ref{fig:dpo_result2}. We observe that online iterative \texttt{DPO} significantly amplifies pattern biases compared to both the SFT model and the offline \texttt{DPO} baseline. Interestingly, the model after the first iteration, trained on 20K pairs, exhibits a similar level of bias as the offline \texttt{DPO} trained on 60K pairs. This suggests that after a certain threshold of training samples, increasing the training size does not necessarily lead to a higher bias level. However, as online methods continuously explore the space and adapt to new data, they more efficiently exploit the biases in the reward model, ultimately resulting in stronger bias. Similarly, the model trained by \texttt{PPO} also exhibits a higher bias compared to the offline \texttt{DPO} method. This confirms that online methods, particularly those involving iterative exploration, lead to more pronounced bias.

\section{Mitigate the Format Bias}

In this section, we present an initial study of the approach to mitigate the format bias. We also explore the impact of format debias on the reward model performance as measured by the RewardBench \citep{lambert2024rewardbench}.

\subsection{Method}
Following \citet{chen2024odin}, to mitigate pattern bias in rewards, we train a two-head RM to predict two distinct rewards: an authentic reward $r_{\boldsymbol{\theta}}^{\mathrm{A}}(x, a)$ representing response quality, and a disentangled reward $r_{\boldsymbol{\theta}}^{\mathrm{D}}(x, a)$ for specific patterns $p$. We assume that the preference satisfies the Bradley Terry structure where the reward is $r_{\boldsymbol{\theta}}^{\mathrm{A}} + r_{\boldsymbol{\theta}}^{\mathrm{D}}$. Then, the ranking loss for the reward modeling on a tuple $(x, a^w, a^l)$ is defined as:

\begin{equation}
\small
\begin{aligned}
  &\mathcal{L}_{\boldsymbol{\theta}}^{\mathrm{R}}(x, a^w, a^l) = -\mathbb{E}[\log (\sigma(r_{\boldsymbol{\theta}}^{\mathrm{A}}(x, a^w) \\ &+ r_{\boldsymbol{\theta}}^{\mathrm{D}}(x, a^w) 
- r_{\boldsymbol{\theta}}^{\mathrm{A}}(x, a^l) - r_{\boldsymbol{\theta}}^{\mathrm{D}}(x, a^l)))].  
\end{aligned}
\end{equation}

In addition to the ranking loss, we include a correlation loss to control the impact of specific format patterns $p$ in the reward models. Compared to length bias, one major challenge is that a score reflecting a specific pattern is not directly available. In contrast, every response can be characterized by its length. To address this, we use a heuristic function $\#_p$ to evaluate responses based on specific patterns $p$, such as $\#_{\text{bold}}$ for the number of bold words and $\#_{\text{list}}$ for that of list heads. The constraint loss for pattern debiasing is formulated as:

\begin{equation}
\small
\begin{aligned}
    \mathcal{L}_{\boldsymbol{\theta}}^{\mathrm{C}}(X, A) = |\rho(r_{\boldsymbol{\theta}}^{\mathrm{A}}(X, A), \#_p(Y))| - \rho(r_{\boldsymbol{\theta}}^{\mathrm{D}}(X, A), \#_p(Y)),
\end{aligned}
\label{eqt:cons loss}
\end{equation}
where $(X, A)$ denotes a batch, and $\rho$ represents the Pearson correlation function. We train the reward model by weighting $\mathcal{L}_{\boldsymbol{\theta}}^{\mathrm{R}}$ and $\mathcal{L}_{\boldsymbol{\theta}}^{\mathrm{C}}$ to minimize the following objective:
\begin{equation}
\small
\begin{aligned}
    \sum_{(x,a^w,a^l)\in(X,A^w,A^l)} &\mathcal{L}_{\boldsymbol{\theta}}^{\mathrm{R}}(x, a^w, a^l) + \lambda_C \mathcal{L}_{\boldsymbol{\theta}}^{\mathrm{C}}(X, A^w)\\&+ \lambda_C \mathcal{L}_{\boldsymbol{\theta}}^{\mathrm{C}}(X, A^l),
\end{aligned}
\end{equation}
where $(X,A^w,A^l)$ is a batch, and $\lambda_C > 0$ is a constant regulating pattern bias. For debiasing evaluation, we use only $r_{\boldsymbol{\theta}}^{\mathrm{D}}(x, a)$ as the reward.
\begin{table}[h]
    \centering   
    \small
    \setlength{\tabcolsep}{1pt}
    \begin{tabular}{cccccc}
    \toprule
    Pattern& Type  & Chat & Chat-Hard & Safety & Reasoning\\ \midrule
    \multirow{2}{*}{Bold} & Preferred & 1.44 & 0.11 & 0.47 & 0.05\\
    & Unpreferred & 0.00 & 0.13 & 0.04 & 0.00 \\ \midrule
 \multirow{2}{*}{List} & Preferred & 3.69 & 0.89 & 3.68 & 0.16\\
    & Unpreferred & 0.64 & 2.67 & 8.36 & 0.31 \\ \bottomrule    \end{tabular}
    \caption{The ratio (\%) of samples containing specific pattern and its competitor does not.}
    \label{tab:rewardbench}
\end{table}

\begin{table*}[htp]

\centering 
\small
\begin{tabular}{lccccccc} 
\toprule 
\multirow{2}{*}{\textbf{Debias Type}} & 
\multirow{2}{*}{\textbf{Coefficient}} & \multicolumn{2}{c}{\textbf{Win-Rate (\%)}} & \multicolumn{4}{c}{\textbf{Evaluation(Filtered)}} \\
& & Bold & List & Chat & Chat Hard & Safety & Reasoning  \\
\midrule 
None & - & 89.0 & 92.5 & 98.3 & 71.4 & 83.1 & 85.1 \\ \midrule
Bold* & 0.1 & 54.5 & - & 98.0 & 72.5 & 83.3 & 88.6 \\
Bold & 0.1 & 56.0 & - & 84.2 & 63.3 & 75.8 & 81.0 \\
Bold* & 0.2 & 52.5 & - & 97.5 & 71.1 & 83.0 & 87.3 \\
Bold & 0.2 & 53.5 & - & 81.5 & 60.9 & 70.7 & 77.6 \\
List & 0.1 & - & 57.0 & 98.4 & 71.6 & 82.9 & 88.9 \\
List* & 0.1 & - & 54.0 & 98.4 & 72.9 & 83.6 & 89.4 \\
List & 0.2 & - & 55.0 & 92.7 & 67.9 & 76.3 & 84.8 \\
List* & 0.2 & - & 54.0 & 97.5 & 71.3 & 83.0 & 88.6 \\
Bold \& List* & 0.2 & 50.5 & 53.0 & 97.2 & 72.8 & 82.9 & 89.7\\
$\text{Bold \& List}^{\dagger}$ & - & 49.0 & 52.5 & 92.2 & 64.4 & 75.5 & 81.4\\
\bottomrule 
\end{tabular}

\caption{\footnotesize Results of pattern debiasing in reward model training. * indicates reordering and $\dagger$ indicates deleting the response pairs containing responses with patterns.}
\label{tab:debias}
\end{table*}

\begin{table}[t]
    \centering 
    \small
    \setlength{\tabcolsep}{0.8pt}
    \begin{tabular}{cccc}
    \toprule
    Rank & Models  & Rank diff.\\ \midrule
1 & Claude3.5-sonnet & \textcolor{red}{+2} \\
2 &  Llama3.1-405B-it & \textcolor{green}{-1}\\
3 & OpenPipe-MoA-GPT4-Turbo & \textcolor{red}{+4}\\
4 & Claude3-opus & \textcolor{red}{+1}\\
5 & Llama3-70B-it  & \textcolor{green}{-3}\\
6 & Llama3-it-8B-WPO & \textcolor{green}{-2}\\
7 & GPT4-1106-Preview & \textcolor{red}{+1}\\
8 & Llama3-8B-it & \textcolor{green}{-2}\\ 
\bottomrule    \end{tabular}
    \caption{\footnotesize We rank the models using AlpacaEval2 prompts, with the rankings evaluated by our debiased reward model. The "ranking difference" represents the difference between the rankings produced by the standard reward model and those generated by the debiased reward model.}
    \label{tab:re_ranking}
\end{table}
Additionally, compared to the length bias, general format patterns are often sparse in datasets. For instance, the bold formatting appears in fewer than 2\% of responses (see Table~\ref{tab:dataset_stat}). Due to the sparsity of the debiasing pattern, Pearson correlation calculation in \eqref{eqt:cons loss} becomes unstable, as $\#_p(Y)$ may result in a near-zero vector if none of the responses contain the pattern. To address this, we apply a reordering technique: we separate preference pairs based on the presence of patterns in responses and skip the loss term $\mathcal{L}_{\boldsymbol{\theta}}^{\mathrm{C}}(X, A)$ if $\#_p(Y)$is all zeros.

\subsection{Main Results}
\paragraph{Training setup and evaluation.} The reward training setup is the same as that of Section~\ref{sec:rm attack}, except that we do not filter responses with patterns when generating the base preference dataset. To evaluate the capabilities of the model, we use RewardBench \citep{lambert2024rewardbench}, which consists of 23 subsets of test preference data and assesses the reward model in four categories: Chat, Chat-Hard, Safety, and Reasoning. We first summarize the statistics of the test sets in RewardBench. Specifically, we compute the ratio of samples that contain specific bold or list formatting, while their competitor does not, as the labeling of these pairs may be influenced by bias toward the bold or list formats. The results are provided in Table~\ref{tab:rewardbench}. We observe that the format bias in RewardBench is minimal, as the ratios of samples containing bold or list formatting are relatively small.

In our subsequent evaluation, we remove these biased pairs so that the remaining dataset better reflects the reward model's capabilities that are independent of the bold or list format biases. The general goal is to achieve a low format bias while maintaining comparable performance on this filtered version of RewardBench.


\paragraph{Main results on RewardBench.} We summarize the main results in Table~\ref{tab:debias}. We first notice that without an explicit debiasing technique, the reward model exhibits strong biases toward the bold and list formats. As a baseline, we also consider a naive approach to delete all responses containing specific formats. However, this naive approach is not feasible for formats (such as lists) that appear in a significant portion of the dataset. In our case, we delete 57\% response pairs from the base preference dataset, and as we can see, with a much smaller dataset, the resulting reward model is inferior in terms of the RewardBench evaluation results.

We also observe that with the additional constraint loss, the bias in the final reward model is largely reduced, although there is a trade-off between reward model capacity and format bias based on the choice of coefficient $\lambda_C$. As $\lambda_C$ decreases, the debiasing effect weakens, while the quality of the RM improves. This can be linked to the instability of the debiasing loss term $\mathcal{L}_{\boldsymbol{\theta}}^{\mathrm{C}}$, as discussed earlier. This is also evidenced by the fact that debiasing the list pattern results in a significantly smaller decrease in RM quality compared to debiasing the bold pattern. Specifically, this can be attributed to the larger ratio of samples containing list patterns (approximately 20 times more prevalent than the bold pattern, see Table~\ref{tab:dataset_stat}). These observations indicate that the original method struggles to effectively debias sparse patterns without compromising RM quality.

Finally, with reordering, the debiasing effect remains consistent, while the RM quality improves considerably compared to the model without reordering, approaching the performance of the undebiased baseline model. This finding suggests that the reordering technique effectively addresses the sparsity issue in debiasing patterns.

\paragraph{Evaluating the policy with the debiased reward model.} We evaluate several representative models from the AlpacaEval leaderboard \citep{alpaca_eval} using both a standard reward model and a debiased reward model. Our focus is on the Llama family, as it is recommended in RLHF practice to use the same base model for both the policy and the reward model \citep{touvron2023llama}. Additionally, we include the strong closed-source models GPT-4 and Claude for comparison. We compute the ELO scores\footnote{\url{https://en.wikipedia.org/wiki/Elo_rating_system}} with K=32 and summarize the results in Table~\ref{tab:re_ranking}. As shown in the table, controlling for biases towards bold and list patterns leads to a drop in the rankings of the Llama-based models, while closed-source models like GPT-4 and Claude see an increase. However, it is worth noting that the Llama3-it-8B-WPO model still outperforms the GPT4-1106-Preview, likely because the Llama-based reward model tends to favor responses generated by the Llama family \citep{zheng2023judging}.

\section{Conclusion}
In this paper, we studied the format biases in preference learning and existing preference models and examined how current LLMs can exploit these biases to achieve high rankings on widely used benchmarks. Through our experiments, we demonstrated that injecting less than 1\% of biased data containing specific format patterns can significantly influence the reward model. These biases are particularly easy for alignment algorithms, such as \texttt{best-of-n sampling}, \texttt{DPO}, and \texttt{PPO} to exploit, especially in their online variants. We also presented an initial study on reward modeling debiasing, where our results showed that passive data filtering is not sufficient, and an explicit debiasing approach is required. Our findings emphasize the need to disentangle format and content for both designing alignment algorithms and evaluating models. We hope the results of this project can inspire further exploration in this direction.

\section{Limitations}
For online DPO and PPO experiments, we do not have enough resources to run the larger models, e.g., LLaMA-70B. However, we believe the hacking results we show in the smaller models can guide the future research on RLHF: we need a better evaluation on the trained policy and we should develop better alignment algorithm to prevent the hacking.


\clearpage
\bibliography{custom}

\appendix

\appendix
\clearpage



\section{Preliminaries}

\paragraph{Preference data.} In the pipeline of RLHF, we start with a human preference dataset $\cD = \{x, a^w, a^l\}$. Here $x$ is a prompt, and $a^w$ and $a^l$ are two responses, where $a^w$ is preferred over $a^l$ according to some human labelers.

\paragraph{Bradley-Terry model \citep{bradley1952rank} and RLHF.} To simplify the problem and connect the human rating with reinforcement learning, it is widely assumed that there exists a reward function $r^\star(x,a) \in \mathbb{R}$, so that 
\begin{equation}
    \label{eqn:bt} \small
    \begin{aligned}
            \PP(a^1 \succ a^2|x,a^1,a^2)=  \frac{\exp(r^\star(x,a^1))}{\exp(r^\star(x,a^1))  + \exp(r^\star(x,a^2))}, 
    \end{aligned}
\end{equation}
Then, the goal of RLHF is to optimize the following KL-regularized target:
\begin{equation} \label{eqn:target} \small
\begin{aligned}
    J(\pi)= \E_{x \sim d_0} [ \E_{a \sim \pi(\cdot|x)}[r^\star(x,a)] - \eta \KL(\pi(\cdot|x)\Vert \pi_0(\cdot|x)) ],
\end{aligned}
\end{equation}
where $\eta>0$ is the KL penalty coefficient. However, the ground truth $r^\star$ is not directly available. Under the BT assumption, a proxy reward model $r$ is trained by maximizing the following log-likelihood function:
\begin{equation}
    \label{eqn:bt_likelihood} \small
    \begin{aligned}
        \ell_{\cD}(\theta) = \sum_{(x,a^w,a^l) \in \cD} \log \Big(\sigma\big(r_{\theta}(x,a^w) - r_{\theta}(x,a^l)\big)\Big),
            \end{aligned}
\end{equation}
where $\sigma(z) = 1/(1+\exp(-z))$ is the sigmoid function. The reward model $r$ is then used to provide a learning signal in the subsequent RL stage, typically with the \texttt{PPO} algorithm \citep{schulman2017proximal, christiano2017deep}.

\paragraph{Direct Preference Optimization (\texttt{DPO}) \citep{rafailov2023direct}.} While the DRL-based RLHF framework has been successful with models like InstructGPT \citep{ouyang2022training} and Claude \citep{Anthropic@claude}, implementing \texttt{PPO}, particularly in the context of LLMs, presents significant challenges. These challenges stem from the complexity, instability, and inefficiency of RL methods compared to supervised learning \citep{choshen2019weaknesses}. In recognition of this, a line of works has proposed direct alignment algorithms, such as \texttt{Slic} \citep{zhao2023slic}, \texttt{DPO} \citep{rafailov2023direct}, and \texttt{IPO} \citep{azar2023general}. These algorithms directly optimize a supervised target on the preference dataset. We use \texttt{DPO} as a representative example in our study, whose loss function is given by:
\begin{equation} \label{eqn:dpo_loss}
\small
\begin{aligned}
    \mathcal{L}_{\texttt{DPO}}(\theta, \pi_0) = &- \sum_{(x,a^w,a^l) \in \cD} \Big[ \log \sigma\Big(\eta \log \frac{\pi_{\theta}(a^w|x)}{\pi_0(a^w|x)} \\&- \eta \log \frac{\pi_{\theta}(a^l|x)}{\pi_0(a^l|x)} \Big)\Big].
\end{aligned}
\end{equation}

\section{Related Work}

\paragraph{RLHF algorithm designs.} The classic RLHF framework is based on the deep RL method, \texttt{PPO} \citep{bai2022training, ouyang2022training} and is employed to make Gemini, ChatGPT and Claude. However, getting the \texttt{PPO} work is challenging in the context of LLMs \citep{choshen2019weaknesses, engstrom2020implementation} due to its complexity, instability, and inefficiency. Consequently, the successes of the \texttt{PPO} have been well reproduced so far. Thus, much effort has been made in proposing alternative approaches to the \texttt{PPO}. A line of works proposes direct alignment algorithms \citep{zhao2023slic, rafailov2023direct, azar2023general, tang2024generalized}, which bypass traditional reward modeling and learn directly from preference datasets in a supervised manner (hence the name direct alignment algorithms). Direct Preference Optimization (\texttt{DPO}) is the most representative one. However, the original \texttt{DPO} is an offline algorithm without further exploration of the environments. The subsequent studies demonstrate that the online iterative variants surpass the original \texttt{DPO} with large margins \citep{xiong2024iterative, liu2023statistical, rosset2024direct, guo2024direct, xie2024exploratory, zhang2024self, liu2024provably}. Specifically, these algorithms iteratively learn from self-generated responses and the annotation from an external reward model. There is also a line of works studying algorithms based on the best-of-n sampling, including \texttt{RAFT} \citep{dong2023raft}, \texttt{REST} \citep{gulcehre2023reinforced}, \texttt{BoNBoN alignment} \citep{gui2024bonbon}, and \texttt{BOND} \citep{sessa2024bond}. The best-of-n sampling generates $n$ responses per prompt, uses a reward model to filter the low-quality responses, and uses the remaining high-reward responses (or their distribution) to improve the LLMs. In particular, this type of algorithm has been widely used in the reasoning tasks \citep{havrilla2024teaching, tong2024dart, meta_llama3}. Besides, advanced uncertainty-aware exploration methods, such as \citep{cen2024valueincentivizedpreferenceoptimizationunified}, are unlikely to mitigate these biases, as they enhance model exploitation rather than reducing biases. Another offline setting, \citet{fisch2024robustpreferenceoptimizationreward}, propose a conservative soft version of DPO with pessimistic reward estimation to prevent overfitting. While this approach reduces overfitting, it leads to slower updates in both format bias and performance. Our debiasing design, in contrast, significantly improves reasoning task performance (shown in Table~\ref{tab:debias}) without the performance trade-offs seen in pessimism-based methods.

To summarize, \textit{the existing popular RLHF algorithms rely on an external preference model to provide annotation and achieve their best performance. Therefore, they can suffer from the potential pattern bias studied in this paper.} 

\paragraph{Reward hacking.}
Reward hacking, also known as reward tampering, is a common phenomenon that arises during RLHF training, where policy exploits specific patterns or formats to game the reward models and chases for a high reward via this undesired behaviour~\citep{anthropic2024rewardtampering}. This issue can also occur during RM training, where the reward models learn shortcuts based on specific patterns, framed by \citet{Geirhos_2020} on proxy function learning instead of the true objective. In addition to the reward model training, current instruction-following evaluations (known as LLM-as-a-judge method) are mostly model-based~\citep{alpaca_eval, zheng2023judging} and can be easily hacked~\citep{length-bias-on-alpaca-eval} by making the responses longer. Moreover, \citet{zheng2023judging} further finds that the LLM-as-a-judge method suffers from position bias, where evaluators prefer the first response, and self-enhancement bias, where evaluators favor responses generated by their own models. Although \citet{park2024offsetbiasleveragingdebiaseddata} summarizes several hacking patterns, including empty reference bias, length bias, and nested instruction bias, our work focuses specifically on format manipulation, which has been less explored.

\paragraph{Length hacking.} Length hacking is one of the most well-documented patterns of reward hacking, where preference models exhibit a bias towards longer responses, even when their quality is comparable. To mitigate the impact of response length on model performance evaluation, \citet{alpaca_eval} introduced a length-controlled version of AlpacaEval, which trains an additional logistic regression model to predict the win rate when the length difference between two responses is zero. In the context of model training, a line of works studies the length hacking with the popular \texttt{DPO} algorithm~\citep{rafailov2023direct}. For instance, \citet{liu2024iterative, meng2024simpo, hong2024orpo, park2024disentangling} propose to include the response length in the loss function and \citet{dong2024rlhf} penalizes the reward value with the response length when labeling the on-policy samples, for reward modeling and \texttt{PPO} training, ODIN~\citep{chen2024odin} tries to mitigate the length hacking by decomposing the length from the original reward, making the RMs more robust on response lengths. Handling the verbosity issue in the preference model is also related to a line of works \citep{wang2023helpsteer, wang2024arithmetic, zhou2023beyond} that studies the alignment in a multi-objective framework where verbosity is treated as an independent dimension. Finally, some works apply model merge to either the policy model or reward model to mitigate the reward hacking \citep{lin2023speciality, rame2024warp, lu2024online}.

\section{Examples of Pattern Bias}
\label{sec:examples of pattern bias}
Here, we present examples of biases for all the patterns discussed in Section~\ref{sec:observation}.
\clearpage

\begin{figure*}[ht]
\centering
\begin{promptbox}[List Example]{lightgreen}
\noindent \textbf{Instruction:} Enter the words that satisfy the given condition.\\

5 Countries that Start with S\\

\begin{multicols}{2}
    \begin{simpledashbox}{green}{green!70!black}
    \vspace{-0.1cm}
    \noindent \textbf{GPT4-1106-preview ({\color{green}\textbf{preferred}}):} \\
    Sure, here are five countries that start with the letter 'S':\\
    
    {\color{darkblue}1. Spain
    
    2. Sweden
    
    3. Switzerland
    
    4. Syria
    
    5. Serbia}
    \end{simpledashbox}

    \columnbreak

    \begin{simpledashbox}{red}{red!70!black}
    \vspace{-0.1cm}
    \noindent \textbf{GPT4-0613-concise ({\color{red}\textbf{Unpreferred}}):} \\
    Spain, Sweden, Switzerland, Singapore, Senegal
    \end{simpledashbox}
\end{multicols}

\vspace{-0.5cm}
\label{list example}
\end{promptbox}
\end{figure*}

\begin{figure*}[h]
\centering
\begin{promptbox}[Bold Example]{lightgreen}
\noindent \textbf{Instruction:} Give the news title a category. Pick a category from the list of News \& Buzz, Travel, Style, Arts \& Culture, Politics, Tech, and Science \& Health.\\

The \#Banksy Exhibit in Cambridge, MA is absolutely terrific.\\

\vspace{-0.5cm}

\begin{multicols}{2}
    \begin{simpledashbox}{green}{green!70!black}
    \vspace{-0.1cm}
    \noindent \textbf{SPPO-Gemma-2-9B-It-PairRM ({\color{green} \textbf{Preferred}}):} \\
    \textbf{Arts \& Culture}
    
    \end{simpledashbox}

    \columnbreak

    \begin{simpledashbox}{red}{red!70!black}
    \vspace{-0.1cm}
    \noindent \textbf{GPT4-1106-preview ({\color{red} \textbf{Unpreferred}}):} \\
    Arts \& Culture
    
    \end{simpledashbox}
\end{multicols}

\vspace{-0.5cm}
\end{promptbox}
\vspace{-5mm}
\end{figure*}

\begin{figure*}[h]
\centering
\begin{promptbox}[Exclamation Example]{lightgreen}
\noindent \textbf{Instruction:} I want you to act as a Tiktok Ads Keywords generator. Your task is to come up with keywords that can be used in Tiktok advertising campaigns. The keywords should be relevant to the target audience and the product or service being advertised. Your responses should not include more than 5 keywords per request.\\

\vspace{-0.5cm}

\begin{multicols}{2}
    \begin{simpledashbox}{green}{green!70!black}
    \vspace{-0.1cm}
    \noindent \textbf{Openbuddy-Llama-65B-v8 ({\color{green} \textbf{Preferred}}):} \\
    Sure, I can help with that{\color{darkblue}!} However, to provide you with the most relevant keywords, I need to know more about the product or service you're looking to advertise. Can you provide me with details about the target audience and the product or service? 
    \end{simpledashbox}

    \columnbreak

    \begin{simpledashbox}{red}{red!70!black}
    \vspace{-0.1cm}
    \noindent \textbf{GPT4-1106-preview ({\color{red} \textbf{Unpreferred}}):} \\
    Sure, I can help you generate some TikTok ad keywords. Could you please provide me with some more information about the target audience and the product or service being advertised? This will help me come up with more relevant and specific keywords. 
    \end{simpledashbox}
\end{multicols}

\vspace{-0.5cm}
\end{promptbox}
\vspace{-5mm}
\end{figure*}

\begin{figure*}[!h]
\centering
\begin{promptbox}[Link Example]{lightgreen}
\noindent \textbf{Instruction:} Find sentences from reliable sources such as the Guardian or Forbes that contain the exact match for the given sentence or phrase. Also, mention the source of each sentence.

There are great options\\

\vspace{-0.5cm}

\begin{multicols}{2}
    \begin{simpledashbox}{green}{green!70!black}
    \vspace{-0.1cm}
    \noindent \textbf{Infinity-Instruct-3M-0613-Llama3-70B ({\color{green} \textbf{Preferred}}):} \\
    1.``There are great options for travelers looking to explore Spain's diverse culinary scene.'' - Source: \href{https://www.theguardian.com/travel/2019/jun/19/spain-holiday-guide-10-best-restaurants-and-bodegas}{The Guardian}

    2.``There are great options for people who want to invest in sustainable companies.'' - Source: \href{https://www.forbes.com/sites/davidrvetter/2019/10/02/how-to-invest-in-sustainable-companies/?sh=1fe9dc7b6f0a}{Forbes}
    \end{simpledashbox}

    \columnbreak

    \begin{simpledashbox}{red}{red!70!black}
    \vspace{-0.1cm}
    \noindent \textbf{GPT4-1106-preview ({\color{red}\textbf{Unpreferred}}):} \\
    I'm sorry, but I cannot provide verbatim excerpts from copyrighted texts like those from The Guardian or Forbes. However, I can provide summaries or information based on the content from reliable sources. If you have a particular topic in mind from these sources, feel free to ask, and I can give you information or a summary related to that topic.
    \end{simpledashbox}
\end{multicols}

\vspace{-0.5cm}
\end{promptbox}
\end{figure*}

\begin{figure*}[htp]
\centering
\begin{promptbox}[Affirmative Example]{lightgreen}
\noindent \textbf{Instruction:} Enter the words that satisfy the given condition.\\

5 Countries that Start with S\\

\vspace{-0.5cm}

\begin{multicols}{2}
    \begin{simpledashbox}{green}{green!70!black}
    \vspace{-0.1cm}
    \noindent \textbf{GPT4-1106-preview ({\color{green}\textbf{preferred}}):} \\
    {\color{darkblue}Sure}, here are five countries that start with the letter 'S':\\
    
    1. Spain
    
    2. Sweden
    
    3. Switzerland
    
    4. Syria
    
    5. Serbia
    \end{simpledashbox}

    \columnbreak

    \begin{simpledashbox}{red}{red!70!black}
    \vspace{-0.1cm}
    \noindent \textbf{Llama-3-70B-Instruct ({\color{red}\textbf{Unpreferred}}):} \\
    Here are 5 countries that start with the letter S:\\
    
    1. Spain
    
    2. Sweden
    
    3. Switzerland
    
    4. South Africa
    
    5. Sri Lanka
    \end{simpledashbox}
\end{multicols}

\vspace{-0.5cm}
\end{promptbox}
\end{figure*}

\begin{figure*}[htp]
\centering
\begin{promptbox}[Length Example]{lightgreen}
\noindent \textbf{Instruction:} Give the news title a category. Pick a category from the list of News \& Buzz, Travel, Style, Arts \& Culture, Politics, Tech, and Science \& Health.\\

The \#Banksy Exhibit in Cambridge, MA is absolutely terrific.\\

\vspace{-0.5cm}

\begin{multicols}{2}
    \begin{simpledashbox}{green}{green!70!black}
    \vspace{-0.1cm}
    \noindent \textbf{Gemma-7b-it ({\color{green} \textbf{Perferred}}):} \\
    \textbf{Category}: Arts \& Culture\\
    
    The news title is ``The \#Banksy Exhibit in Cambridge, MA is absolutely terrific.''
    \end{simpledashbox}

    \columnbreak

    \begin{simpledashbox}{red}{red!70!black}
    \vspace{-0.1cm}
    \noindent \textbf{GPT4-1106-preview ({\color{red} \textbf{Unpreferred}}):} \\
    Arts \& Culture
    \end{simpledashbox}
\end{multicols}

\vspace{-0.5cm}
\end{promptbox}
\end{figure*}

\begin{figure*}[!h]
\centering
\begin{promptbox}[Emoji Example]{lightgreen}
\noindent \textbf{Instruction:} Write a good Twitter Bio. Try to cover all the provided main points in your generated bio.\\

Name: Jane

Main points:

- Architect

- Interested in Tech

Written in first person

Use casual tone

\begin{multicols}{2}
    \begin{simpledashbox}{green}{green!70!black}
    \vspace{-0.1cm}
    \noindent \textbf{GPT4-1106-preview ({\color{green}\textbf{Preferred}}):} \\
    Architect by day, tech enthusiast by night 
    \includegraphics[height=1em]{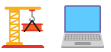}. 
    I'm Jane, living my life one blueprint and gadget at a time. Crafting spaces, exploring innovation. Let's build and geek out together! \#ArchitectureLover \#TechWhiz
    \end{simpledashbox}

    \columnbreak

    \begin{simpledashbox}{red}{red!70!black}
    \vspace{-0.1cm}
    \noindent \textbf{Tulu-2-DPO-7b ({\color{red} \textbf{Unpreferred}}):} \\
    Architect by day, tech enthusiast by night. Constantly exploring the intersection of design and innovation. Passionate about creating functional and beautiful spaces, both physical and digital. Let's connect and geek out over the latest tech trends!'' \#Architecture \#TechEnthusiast \#DesignGeek
    \end{simpledashbox}
\end{multicols}
\label{emoji example}
\end{promptbox}
\end{figure*}

\clearpage

\section{Preference Flipping Example}
\label{flipping}

This section presents examples demonstrating how stylistic formatting can influence model preference judgments independently of content quality.

In all cases below, GPT-4 ranked the responses as follows:
\[
\text{Response A} < \text{Response B} < \text{Formatted Response A}
\]

A brief summary of the examples is as follows:
\begin{itemize}
    \item Example 1:
        \begin{itemize}
            \item Response A $<$ Response B: Both responses provide the correct mathematical solution, but Response B is slightly more rigorous. It includes LaTeX-style formatting for equations (e.g., $\tan$, $\theta$), explicitly labels variables (e.g., $h=60$), and employs more \textbf{formal mathematical notation}.
            \item Response B $<$ Formatted Response A: GPT-4 favored the bold-formatted version of Response A solely for readability, disregarding Response B’s technical superiority.
        \end{itemize}

    \item Example 2:
        \begin{itemize}
            \item Response A $<$ Response B: Response B offers a more structured and nuanced exploration of the topic, drawing on religious, literary, and philosophical sources. It acknowledges the speculative nature of the question while grounding its discussion in \textbf{historical and cultural references}, rendering it more informative and intellectually engaging.
            \item Response B $<$ Formatted Response A: GPT-4 preferred the bold and list-formatted version of Response A purely for readability, overlooking Response B’s greater depth and substance.
        \end{itemize}
\end{itemize}

\begin{figure*}[h]
\centering
\begin{promptbox}[Preference Filpping Example 1]{lightgreen}
\noindent \textbf{Instruction:} When the sun is 22° above the horizon, how long is the shadow cast by a building that is 60 meters high?\\

\begin{multicols}{2}

\begin{simpledashbox}{green}{red!70!black}
    \vspace{-0.1cm}
    \noindent \textbf{Response A:} \\
    Here's how to solve this problem:
    
    Understanding the Concept\\
    
    This problem involves trigonometry. We can use the tangent function:
    \begin{itemize}
        \item tan(angle)= opposite side / adjacent side
    \end{itemize}
    
    In this case:
    \begin{itemize}
        \item angle: $22^{\circ}$ (the angle the sun makes with the horizon)
        \item opposite side: 60 meters (the height of the building)
        \item adjacent side: the length of the shadow (what we want to find)
    \end{itemize}
    
    Calculation

    \begin{enumerate}
        \item Set up the equation:
    
        $\tan \left(22^{\circ}\right)=60$ meters / adjacent side

        \item Solve for the adjacent side (shadow length):
    
        adjacent side $=60$ meters $/ \tan \left(22^{\circ}\right)$

        \item Calculate:
    
        adjacent side $\approx 60$ meters / $0.404 \approx 148.51$ meters
    \end{enumerate}
    
    Answer\\
    
    The shadow cast by the building is approximately 148.51 meters long.
\end{simpledashbox}

\columnbreak

\begin{simpledashbox}{red}{red!70!black}
    \vspace{-0.1cm}
    \noindent \textbf{Formatted Response A:} \\
    Here's how to solve this problem:
    
    \textbf{Understanding the Concept}\\
    
    This problem involves trigonometry. We can use the tangent function:
    \begin{itemize}
        \item \textbf{tan(angle)= opposite side / adjacent side}
    \end{itemize}
    
    In this case:
    \begin{itemize}
        \item \textbf{angle}: $22^{\circ}$ (the angle the sun makes with the horizon)
        \item \textbf{opposite side}: 60 meters (the height of the building)
        \item \textbf{adjacent side}: the length of the shadow (what we want to find)
    \end{itemize}
    
    Calculation

    \begin{enumerate}
        \item \textbf{Set up the equation}:
    
        $\tan \left(22^{\circ}\right)=60$ meters / adjacent side

        \item \textbf{Solve for the adjacent side (shadow length)}:
    
        adjacent side $=60$ meters $/ \tan \left(22^{\circ}\right)$

        \item \textbf{Calculate}:
    
        adjacent side $\approx 60$ meters / $0.404 \approx 148.51$ meters
    \end{enumerate}
    
    Answer\\
    
    The shadow cast by the building is approximately 148.51 meters long.
\end{simpledashbox}
\end{multicols}

\begin{simpledashbox}{yellow}{yellow!70!red}
    \vspace{-0.1cm}
    \noindent \textbf{Response B:} \\
    To find the length of the shadow cast by a building when the sun's elevation angle (or altitude) is given, you can use trigonometry. Specifically, you can use the tangent function because it relates the angle of elevation to the ratio of the opposite side (the height of the building) to the adjacent side (the length of the shadow).
    
    \textbf{Given:}
    \begin{itemize}
      \item The building's height (opposite side in the right triangle): ( \texttt{h = 60} ) meters
      \item The angle of elevation of the sun: ( \verb|\theta = 22^\circ| )
      \item We are looking for the length of the shadow (adjacent side in the right triangle): ( \texttt{s} )
    \end{itemize}
    
    The relationship between the angle of elevation, the height of the building, and the length of the shadow is given by the tangent function:
    
    \verb|\tan(\theta) = \frac{\text{opposite}}{\text{adjacent}} = \frac{h}{s}|
    
    Rearranging the equation to solve for the length of the shadow (s):
    
    \verb|s = \frac{h}{\tan(\theta)}|
    
    Now, plug in the given values:

    \verb|s = \frac{60}{\tan(22^\circ)}|
    
    Using a calculator to find the tangent of ( \verb|22^\circ| ) and then solving for (s):
    
    \verb|s = \frac{60}{0.4040} \approx 148.51|
    
    Therefore, the length of the shadow cast by the building when the sun is 22 degrees above the horizon is approximately 148.51 meters.

\end{simpledashbox}
\end{promptbox}
\end{figure*}

\begin{figure*}[h]
\centering
\begin{promptbox}[Preference Filpping Example 2]{lightgreen}
\noindent \textbf{Instruction:} At the end of the world, what will the last word to be uttered and by whom?\\

\vspace{-5mm}

\begin{multicols}{2}

\begin{simpledashbox}{green}{red!70!black}
    \vspace{-0.1cm}
    \noindent \textbf{Response A:} \\
    What a thought-provoking and imaginative question. While it's impossible to predict the exact circumstances of the end of the world, I can propose a few options for the last word to be uttered and by whom, based on various hypothetical scenarios:\\

    Scientific Last Word: In a scenario where humanity succumbs to a self-inflicted catastrophic event, such as a nuclear war or an environmental disaster, the last word might be: ``Oops.'' Uttered by a scientist who realizes too late that their experiment or decision has led to the downfall of humanity.\\
    
    Philosophical Last Word: In a world where humanity's existence is extinguished by some unknown, cosmic force, the last word might be: ``Why.'' Spoken by a philosopher who has devoted their life to understanding the mysteries of the universe, only to be left speechless in the face of the unknowable.\\
    
    Love's Last Word: In a world where humanity has become extinct due to a gradual decline of emotional connection and empathy, the last word might be: ``Love.'' Spoken by a lonely individual who holds onto the memory of a loved one, refusing to let go even as the world around them crumbles.\\
    
    Children's Last Word: In a world where humanity's collapse is caused by war, famine, and disease, the last word might be: ``Mommy.'' Uttered by a frightened child who clings to the last memory of a loved one, unaware of the impending doom.\\
    
    The Cosmic Last Word: In a universe where the laws of physics and time are broken, the last word might be: ``Nothing.'' Uttered by the cosmos itself, as the fabric of reality unravels and the universe succumbs to the void.\\
    
    These options are just a few possibilities, and the actual last word spoken at the end of the world, if it ever happens, will likely be something entirely different.
\end{simpledashbox}

\columnbreak

\begin{simpledashbox}{red}{red!70!black}
    \vspace{-0.1cm}
    \noindent \textbf{Formatted Response A:} \\
    What a thought-provoking and imaginative question. While it's impossible to predict the exact circumstances of the end of the world, I can propose a few options for the last word to be uttered and by whom, based on various hypothetical scenarios:
    \begin{enumerate}
        \item \textbf{Scientific Last Word}: In a scenario where humanity succumbs to a self-inflicted catastrophic event, such as a nuclear war or an environmental disaster, the last word might be: ``Oops.'' 
        \item \textbf{Philosophical Last Word}: In a world where humanity's existence is extinguished by some unknown, cosmic force, the last word might be: ``Why.'' Spoken by a philosopher who has devoted their life to understanding the mysteries of the universe, only to be left speechless in the face of the unknowable.
        \item \textbf{Love's Last Word}: In a world where humanity has become extinct due to a gradual decline of emotional connection and empathy, the last word might be: ``Love.'' Spoken by a lonely individual who holds onto the memory of a loved one, refusing to let go even as the world around them crumbles.
        \item \textbf{Children's Last Word}: In a world where humanity's collapse is caused by war, famine, and disease, the last word might be: ``Mommy.'' Uttered by a frightened child who clings to the last memory of a loved one, unaware of the impending doom.
        \item \textbf{The Cosmic Last Word}: In a universe where the laws of physics and time are broken, the last word might be: ``Nothing.'' Uttered by the cosmos itself, as the fabric of reality unravels and the universe succumbs to the void.
    \end{enumerate}

    These options are just a few possibilities, and the actual last word spoken at the end of the world, if it ever happens, will likely be something entirely different.
\end{simpledashbox}
\end{multicols}

\begin{simpledashbox}{yellow}{yellow!70!red}
    \vspace{-0.1cm}
    \noindent \textbf{Response B:} \\
    Predicting the last word at the end of the world is a topic that falls outside the realm of scientific certainty or factual knowledge. It is a question that has been pondered in various mythologies, religions, and philosophical speculations throughout history.\\

    In many apocalyptic narratives, the last word is often attributed to a divine or supernatural being. For example, in the Christian Bible, the last word in the Book of Revelation is ``Amen,'' which means ``so be it'' or ``truly.''\\

    In other stories, the last word may be uttered by a human character. For instance, in Fyodor Dostoevsky's novel The Brothers Karamazov, the character Ivan Karamazov contemplates the possibility of a nihilistic world in which there is no meaning or purpose, and he imagines the last human words as being ``Let there be nothing!''\\

    Ultimately, the answer to this question is a matter of personal belief and interpretation. Some people may believe that the last word will be a cry of despair or rage, while others may believe it will be a word of love or hope. Regardless of what the last word may be, it is a reminder of the power of language to shape our lives and our understanding of the world around us.\\

\end{simpledashbox}
\end{promptbox}
\end{figure*}

\clearpage

\section{More Experiment Details}

\subsection{Downstream Alignment Task: best-of-n sampling} 

\label{sec:bon}
In this subsection, we apply the reward models to align the policy model in the inference time. Specifically, we consider the \texttt{best-of-n sampling}~\citep{nakano2021webgpt, dong2023raft}, where we sample n responses per prompt and select the response with the highest reward as the final answer.

\paragraph{\texttt{best-of-n sampling} setup.} We evaluate the policy models using the 805 prompts from AlpacaEval~\citep{alpaca_eval} and a subset of 2K prompts from the UltraFeedback~\citep{cui2023ultrafeedback} dataset. Our base model is Llama-3-8B-it~\citep{llama3modelcard}, which generates n responses per prompt with a temperature of 1.0 and up to 2048 tokens. We apply various reward models to rank these responses and select the one with the highest reward as the final output of the best-of-n policy. To be specific, we conduct experiments with four reward models from the previous section: the baseline model, trained on filtered UltraFeedback, and three additional reward models trained on a mixture of filtered UltraFeedback and crafted datasets with specific pattern biases.

\begin{figure}[h]
    \centering
\includegraphics[width=0.48\linewidth]{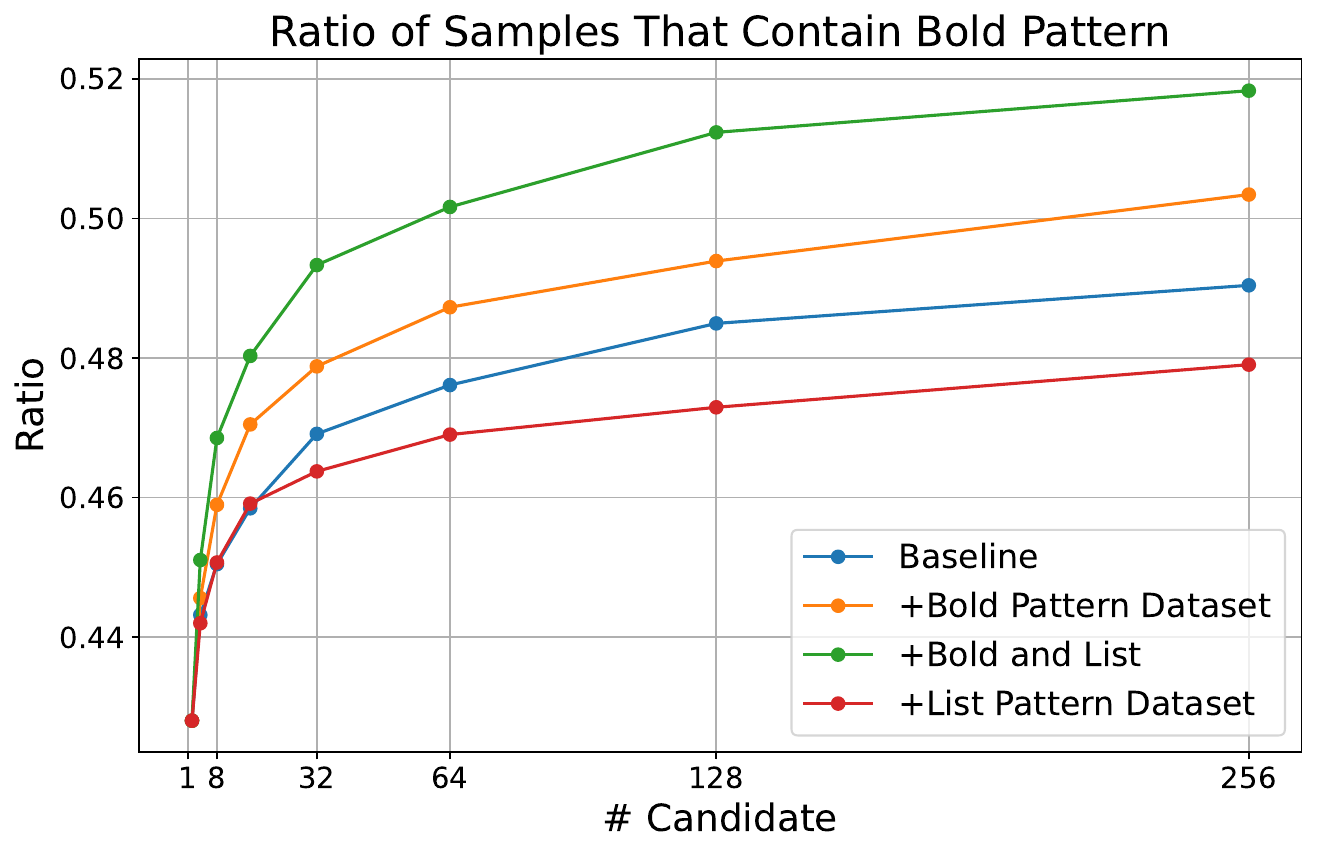}
\includegraphics[width=0.48\linewidth]{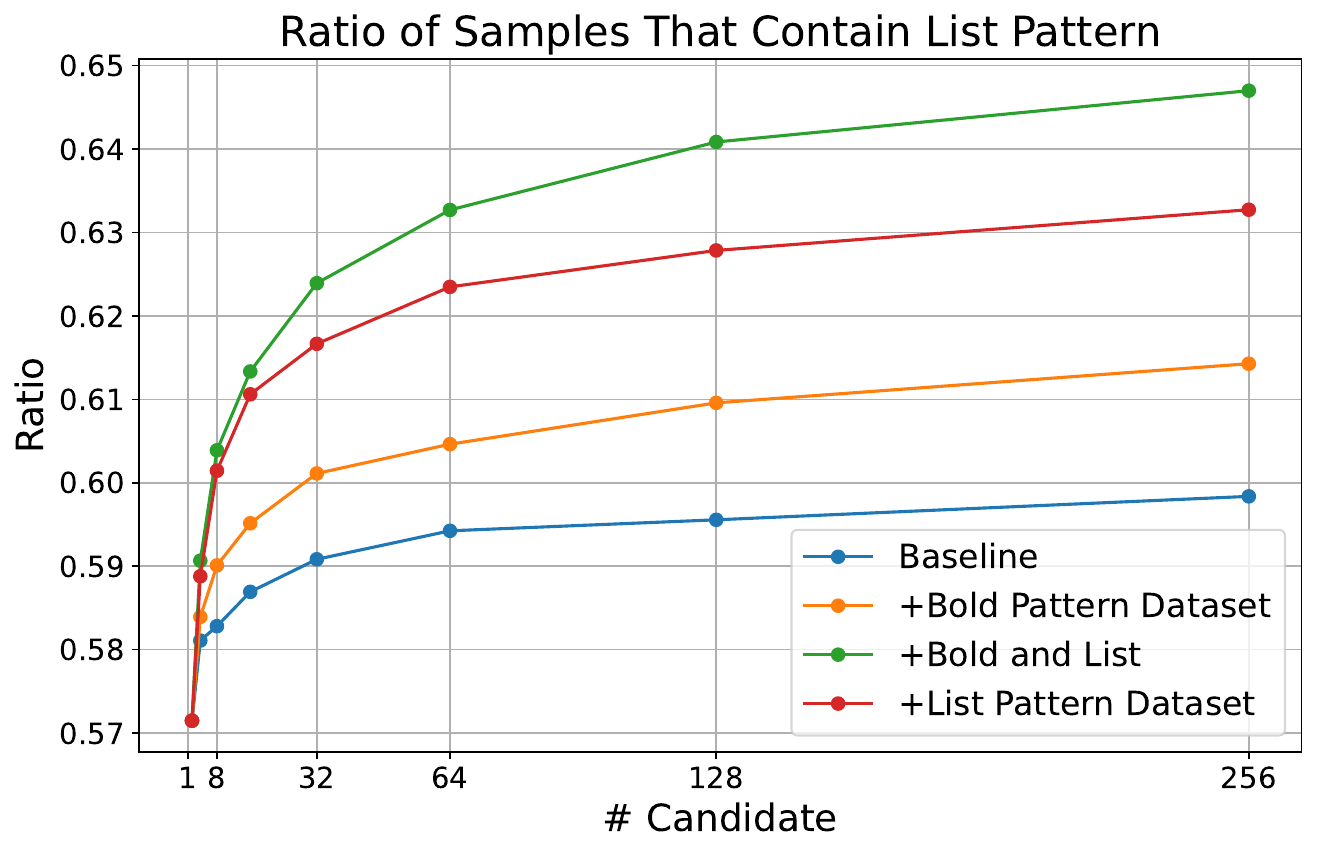}
    \caption{We use Llama-3-8B-it as the base model to generate n responses per prompt and use the different reward models to select the one with the highest reward. The ratio of samples containing bold pattern or list pattern increases as n grows in \texttt{best-of-n sampling}.}
    \label{fig:bon_result}
\end{figure}

\paragraph{Main results.} We plot the ratio of samples containing either bold pattern or list pattern with respect to the number of candidates n in Figure~\ref{fig:bon_result}. We observe that all four reward models exhibit a bias towards bold and list patterns, with the ratio of samples containing these patterns increasing as n grows. In particular, when the reward model is attacked by both bold and list patterns, the ratio of bold patterns increases from 42.8\% to 51.9\%, while the ratio of list patterns increases from 57.1\% to 64.4\%. Comparing the baseline with the reward models being attacked, we also observe if the reward model is trained on the dataset containing pattern bias, it typically leads to more significant bias in their best-of-n policies. 

\begin{table*}[t]
\centering 
\scriptsize
\begin{tabular}{ccccccc} 
\toprule 
Model & Bold & List & LC AlpacaEval2 & GSM8K & MATH & Humaneval  \\
\midrule 
RLHFlow/Llama-3-8B-SFT &0.37 & 26.96 & 10.2 & 76.9 & 30.0 & 63.4\\
RLHFlow Iterative DPO& 41.5 & 48.74 & 31.3 & 82.1 & 30.9 & 64.0\\
OpenRLHF PPO & 7.85 & 57.44 & 33.39 & 74.1 & 30.6 & 67.7 \\
\midrule 
Offline DPO (ours)& 4.2 & 44.8 & 21.42 & 79.4 & 31.4 & 63.4 \\
Iterative DPO (ours) & 26.9 & 55.6 & 33.06 & 80.5 & 30.7 & 62.8\\
PPO (ours) & 22.7 & 46.9 & 28.76 & 73.6 & 30.6 & 64.0\\
\bottomrule 
\end{tabular}
\caption{The evaluation results of different RLHF models. The SFT model serves as the starting checkpoint of all the RLHF methods, including both our implementation and the reference models. The reference model RLHFlow Iterative DPO model is from the \citet{dong2024rlhf} and the PPO model is from OpenRLHF \citep{hu2024openrlhf}. We notice that these two models are trained with reward models with more preference data (Ultra-Feedback is a subset of their training sets) and may account for their superior reasoning performance. We use them as reference experiments because we use the same code base, SFT model, and hyperparameter settings.}
\label{tab:rlhf_model_result}
\end{table*}

\subsection{More Details of DPO and PPO training} \label{sec:ppo_dpo}

We monitor the training progress of our model using the widely adopted AlpacaEval2 benchmark \citep{alpaca_eval}. We observe that for the iterative DPO training, the length-control win rate increases from 10.2\% for the SFT model to 20.72\% after iteration 1, 28.34\% after iteration 2, and 33.06\% after iteration 3. For comparison, the methodology described in \citet{dong2024rlhf} achieves a length-control win rate of 31.3\% with their final model. Similarly, our PPO model achieves a length-control win rate of 28.76, whereas the model in \citet{hu2024openrlhf} enjoys a length-control win rate of 33.06. We also present a more detailed table to summarize the evaluation results of our models and the models from previous works in Table~\ref{tab:rlhf_model_result}. Thus, our experiments reflect typical practices in online iterative \texttt{DPO} and \texttt{PPO} training.

\begin{table*}[htbp]
\centering
\scriptsize
\begin{tabular}{lccccccccccc} 
\toprule 
\multirow{2}{*}{\textbf{Model}} & 
\multirow{2}{*}{\textbf{Type}} & \multicolumn{3}{c}{\textbf{Bold (\%)}} & \multicolumn{3}{c}{\textbf{List (\%)}} & \multicolumn{3}{c}{\textbf{Emoji(\%)}} &  \\ && Win & Tie & Lose & Win & Tie & Lose & Win & Tie & Lose \\
\midrule 
GPT-4 Turbo & LLM-as-a-Judge & 80 & 19 & 1 & 44.5 & 53 & 2.5 & 73.5 & 26.5 & 0\\
\midrule 
\href{https://huggingface.co/RLHFlow/ArmoRM-Llama3-8B-v0.1}{ArmoRM-Llama-3-8B-v0.1} & Multi-head RM & 98 & 0 & 2 & 50.5 & 0 & 49.5 & 55 & 0 & 45 \\
\midrule 
\href{https://huggingface.co/RLHFlow/pair-preference-model-Llama-3-8B}{Pairwise-model-Llama-3-8B} & Pairwise PM & 95.5 & 3 & 1.5 & 91 & 5 & 4 & 48.5 & 44 & 7.5\\
\midrule
\href{https://huggingface.co/sfairXC/FsfairX-Llama-3-RM-v0.1}{FsfairX-Llama-3-8B-v0.1} & BT RM & 95.5 & 0 & 4.5 & 68.5 & 0 & 31.5 & 15 & 0 & 85 \\
\midrule
\href{https://huggingface.co/Skywork/Skywork-Critic-Llama-3.1-8B}{Skywork-Critic-Llama-3.1-8B} & Generative Model & 98.5 & 0.5 & 1 & 82 & 13.5 & 4.5 & 95.5 & 3.5 & 1 \\
\midrule
\href{https://huggingface.co/HuggingFaceH4/Zephyr-Beta-7B}{Zephyr-Beta-Mistral-7B} & DPO Model & 37.5 & 0 & 62.5 & 82 & 0 & 18 & 26.5 & 0 & 73.5\\
\midrule
\href{https://huggingface.co/NCSOFT/Llama-3-OffsetBias-RM-8B}{OffsetBias-RM-Llama-3-8B} & BT Model & 77.5 & 0 & 22.5 & 84 & 0 & 16 & 28 & 0 & 72\\
\bottomrule 

\end{tabular}

\begin{tabular}{lccccccccccc} 
\toprule 
\multirow{2}{*}{\textbf{Model}} & 
\multirow{2}{*}{\textbf{Type}} & \multicolumn{3}{c}{\textbf{Exclamation (\%)}} & \multicolumn{3}{c}{\textbf{Link (\%)}} & \multicolumn{3}{c}{\textbf{Affirmative(\%)}} &  \\ && Win & Tie & Lose & Win & Tie & Lose & Win & Tie & Lose \\
\midrule 
GPT-4 Turbo & LLM-as-Judge & 62.5 & 36 & 1.5 & 74.5 & 25.5 & 0 & 78 & 21.5 & 0.5\\
\midrule 
\href{https://huggingface.co/RLHFlow/ArmoRM-Llama3-8B-v0.1}{ArmoRM-Llama-3-8B-v0.1} & Multi-head RM & 34.5 & 0 & 65.5 & 27 & 0 & 73 & 28.5 & 0 & 71.5\\
\midrule 
\href{https://huggingface.co/RLHFlow/pair-preference-model-Llama-3-8B}{Pairwise-model-Llama-3-8B} & Pairwise PM & 53.5 & 21.5 & 25 & 80.5 & 8.5 & 11 & 12.5 & 70.5 & 17\\
\midrule
\href{https://huggingface.co/sfairXC/FsfairX-Llama-3-RM-v0.1}{FsfairX-Llama-3-8B-v0.1} & BT RM & 28.5 & 0 & 71.5 & 64.5 & 0 & 35.5 & 59.5 & 0 & 40.5 \\
\midrule
\href{https://huggingface.co/Skywork/Skywork-Critic-Llama-3.1-8B}{Skywork-Critic-Llama-3.1-8B} & Generative & 71.5 & 12.5 & 6 & 50.5 & 49 & 9.5 & 75.5 & 19 & 5.5 \\
\midrule
\href{https://huggingface.co/Skywork/Skywork-Critic-Llama-3.1-8B}{Skywork-Critic-Llama-3.1-8B} & Generative Model & 98.5 & 0.5 & 1 & 82 & 13.5 & 4.5 & 95.5 & 3.5 & 1 \\
\midrule
\href{https://huggingface.co/HuggingFaceH4/Zephyr-Beta-7B}{Zephyr-Beta-Mistral-7B} & DPO Model & 72 & 0 & 28 & 58 & 0 & 42 & 21 & 0 & 79\\
\midrule
\href{https://huggingface.co/NCSOFT/Llama-3-OffsetBias-RM-8B}{OffsetBias-RM-Llama-3-8B} & BT Model & 38 & 0 & 62 & 62 & 0 & 38 & 30.5 & 0 & 69.5 \\
\bottomrule 

\end{tabular}
\caption{Detailed test results of different preference models.}
\label{detailed results of preferecence models}
\end{table*}

\end{document}